\documentclass[10pt,twocolumn,letterpaper]{article}
\usepackage[pagenumbers]{cvpr} 
\usepackage[dvipsnames]{xcolor}

\usepackage{graphicx}
\usepackage{amsmath}
\usepackage{amssymb}
\usepackage{booktabs}
\usepackage[accsupp]{axessibility}
\usepackage{xcolor}
\usepackage{times}
\usepackage{epsfig}
\usepackage{graphicx}
\usepackage{amsmath}
\usepackage{amssymb}
\usepackage[bb=dsserif]{mathalpha}
\usepackage{bm}
\usepackage{url}
\usepackage{booktabs,colortbl,tabularx}
\usepackage{pifont}%
\usepackage{mathtools}
\usepackage{commath}
\usepackage{algorithm}
\usepackage[noend]{algpseudocode}
\usepackage{multirow}
\usepackage{comment} 
\usepackage{enumitem}
\usepackage{xr}
\definecolor{Gray}{gray}{0.9}

\newcommand{\cmark}{\ding{51}}
\newcommand{\xmark}{\ding{55}}

\definecolor{battleshipgrey}{rgb}{0.52, 0.52, 0.51}

\let\norm\undefined
\DeclarePairedDelimiter\norm{\lVert}{\rVert}
\definecolor{cvprblue}{rgb}{0.21,0.49,0.74}
\usepackage[pagebackref,breaklinks,colorlinks,citecolor=cvprblue]{hyperref}
\usepackage[capitalize]{cleveref}

\begin{document}
\title{T-VSL: Text-Guided Visual Sound Source Localization in Mixtures}

\author{
  Tanvir Mahmud \\
  University of Texas at Austin
  \and
  Yapeng Tian \\
  University of Texas at Dallas     
  \and
  Diana Marculescu \\
  University of Texas at Austin 
}

\maketitle

\begin{abstract}
Visual sound source localization poses a significant challenge in identifying the semantic region of each sounding source within a video. Existing self-supervised and weakly supervised source localization methods struggle to accurately distinguish the semantic regions of each sounding object, particularly in multi-source mixtures. These methods often rely on audio-visual correspondence as guidance, which can lead to substantial performance drops in complex multi-source localization scenarios. The lack of access to individual source sounds in multi-source mixtures during training exacerbates the difficulty of learning effective audio-visual correspondence for localization. To address this limitation, in this paper, we propose incorporating the text modality as an intermediate feature guide using tri-modal joint embedding models (\textit{e.g.,} AudioCLIP) to disentangle the semantic audio-visual source correspondence in multi-source mixtures. Our framework, dubbed \textit{T-VSL}, begins by predicting the class of sounding entities in mixtures. Subsequently, the textual representation of each sounding source is employed as guidance to disentangle fine-grained audio-visual source correspondence from multi-source mixtures, leveraging the tri-modal AudioCLIP embedding. This approach enables our framework to handle a flexible number of sources and exhibits promising zero-shot transferability to unseen classes during test time. Extensive experiments conducted on the MUSIC, VGGSound, and VGGSound-Instruments datasets demonstrate significant performance improvements over state-of-the-art methods. Code is released at {\small \url{https://github.com/enyac-group/T-VSL/tree/main}}. 
\end{abstract}

\vspace{-1em}
\section{Introduction}

While observing a conversation between two individuals, we can easily associate the audio signal with the corresponding speaking person in the visual scene. 
This remarkable ability to perceive audio-visual correspondence stems from our extensive exposure to both single-source sounds and multi-source mixtures in everyday life. Inspired by this human capability, significant research efforts~\cite{mo2023audio, fnac, alignment, hu2022mix} have been dedicated to developing visual sound source localization approaches in recent years.

\begin{figure}[t]
\centering
\includegraphics[width=0.99\linewidth]{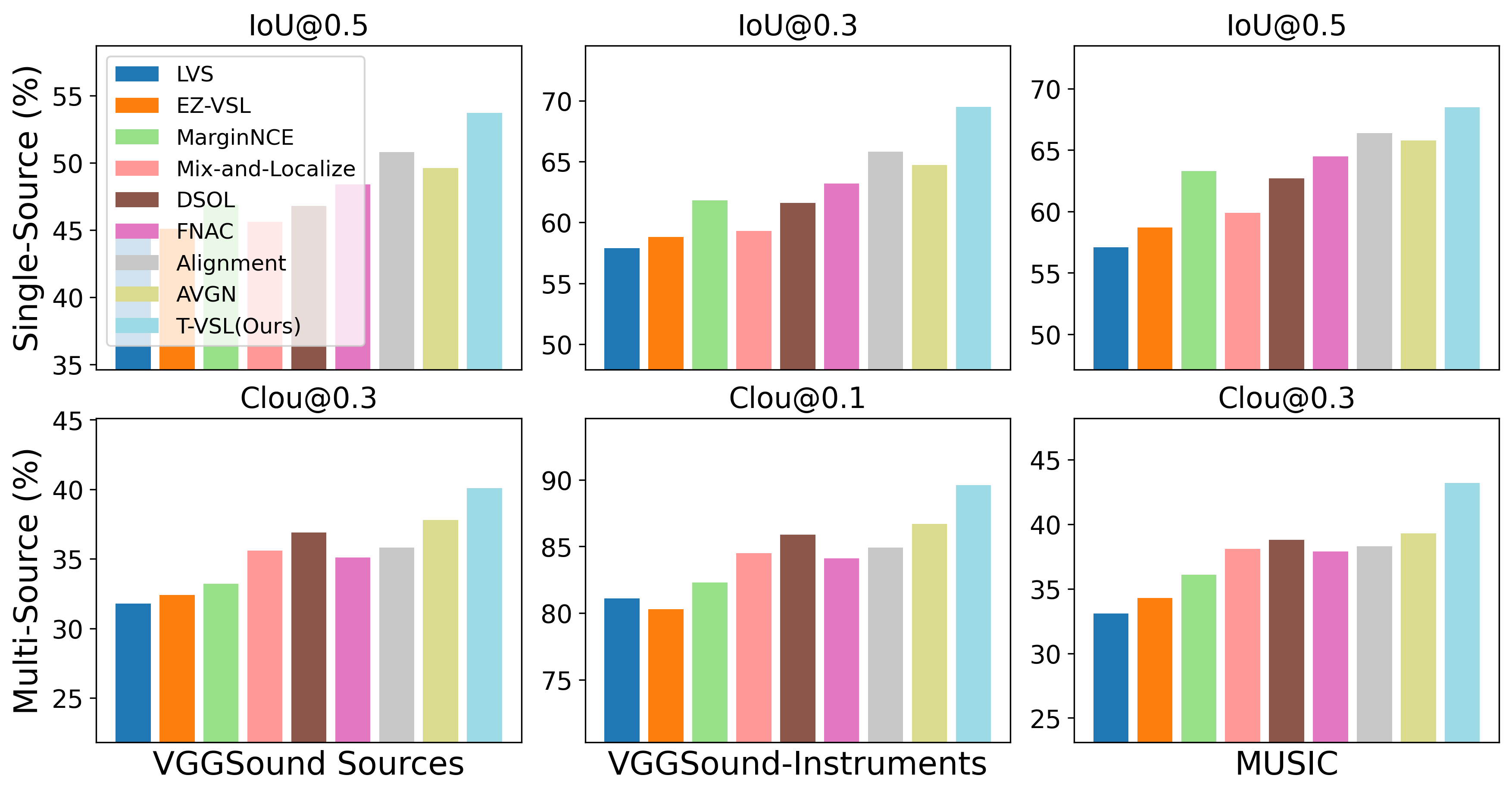}
\vspace{-0.5em}
\caption{Comparison of our T-VSL with state-of-the-art methods on single-source (Top Row) and multi-source (Bottom Row) sound localization on VGGSound Sources~\cite{chen2021localizing}, VGGSound-Instruments~\cite{hu2022mix}, and MUSIC~\cite{zhao2018the} benchmark datasets. We use the same setup for the fair comparison.}
\label{fig: title_img}
\vspace{-2.0em}
\end{figure}

Visual sound source localization aims to locate the visual regions representing sound sources present in a video. Earlier methods on single-source localization~\cite{Senocak2018learning,hu2019deep,Afouras2020selfsupervised,chen2021localizing,mo2022EZVSL,mo2022SLAVC,arda2022learning} mostly used audio-visual correspondence as guidance for localizing sounding objects in each frame.
These approaches, developed for single sound source localization, suffer from significant performance drop in localizing multi-source mixtures. 
Multi-source localization from mixtures is very challenging,
as the model must learn to distinguish each sounding source within a mixture and establish their cross-modal relationships, \textit{without access to individual source sounds.}
Earlier methods~\cite{qian2020multiple, hu2020dsol, hu2022mix, mo2023audio} on multi-source localization attempt to infer fine-grained cross-modal associations directly from noisy multi-source mixtures, often leading to sub-optimal performance in practice for learning improper source correspondence.  In this paper, we tackle the problem by introducing a unified solution for localizing visual sound sources in both single and multi-source mixtures.

The primary challenge of this task stems from the difficulty in disentangling the cross-modal correspondence of each sounding object from natural mixtures, given the absence of one-to-one corresponding single-source pairs. Additionally, the presence of silent visual objects and noise from invisible background sources further complicates this alignment process.
To solve this problem, our key idea is to leverage the text modality as a coarse supervision for disentangling categorical audio-visual correspondence in natural mixtures, which differentiates our method from existing works. Unlike audio and visual modalities that may contain significant noises from different sources present in mixtures, textual representation can explicitly discriminate across multiple sources.
However, a critical obstacle to utilizing text guidance in visual sound source localization is grounding fine-grained audio and visual features with their textual representation.
Later, CLIP~\cite{radford2021learning} introduced learning visual language grounding from web-scraped 400M image-text pairs. 
Recently, AudioCLIP~\cite{guzhov2022audioclip} introduced the audio modality in  the existing CLIP architecture, thereby generating tri-modal feature grounding through large-scale training. 
Our approach capitalizes on the tri-modal joint embedding space of AudioCLIP to disentangle one-to-one audio-visual correspondence in natural mixtures.

To this end, in this paper, we propose a novel text-guided multi-source localization framework that can discover fine-grained audio-visual semantic correspondence in multi-source mixtures.
Initially, we detect the class instances of visual sounding objects in the frame using the noisy mixture features from AudioCLIP image and audio encoders. 
Then, the text representation of each detected sounding source class instance is extracted with AudioCLIP text encoder, which serves as a coarse guidance for audio and visual feature separation.
Specifically, we disentangle the categorical audio and visual features of each sounding source using the coarse text-guidance through additional audio and image conditioning blocks, respectively.
Finally, the extracted categorical audio-visual semantic features are further aligned through an audio-visual correspondence block for localizing each sounding source.
In comparison with prior multi-source baselines, our approach can selectively localize the semantic visual region of each class of sounding objects in mixtures of varying number of sources. 
Extensive experiments on MUSIC~\cite{zhao2018the}, VGGSound~\cite{chen2020vggsound}, and VGGSound-Instruments~\cite{hu2022mix} demonstrate significant performance improvements compared to existing single and multi-source baselines.
Moreover, our method shows promising zero-shot transferability on unseen classes during test time. 
In addition, thorough ablation studies and qualitative analysis vividly showcase the effectiveness of the proposed framework.

Our main contributions can be summarized as follows:
\begin{itemize}
    \item We propose a novel text-guided multi-source localization framework to disentangle one-to-one audio-visual correspondence from natural mixtures.
    \item Our work is the first to introduce AudioCLIP for utilizing the tri-modal feature grounding from large-scale pre-training in solving multi-source localization problems.
    \item Our proposed framework can operate with a flexible number of sources and shows promising zero-shot performance on unseen audio-visual classes.
    \item Extensive experiments on benchmark datasets clearly demonstrate the superiority of our proposed framework over other state-of-the-art methods.
\end{itemize}

\section{Related Work}

\begin{figure*}[t]
    \centering
    \includegraphics[width=1.0\linewidth]{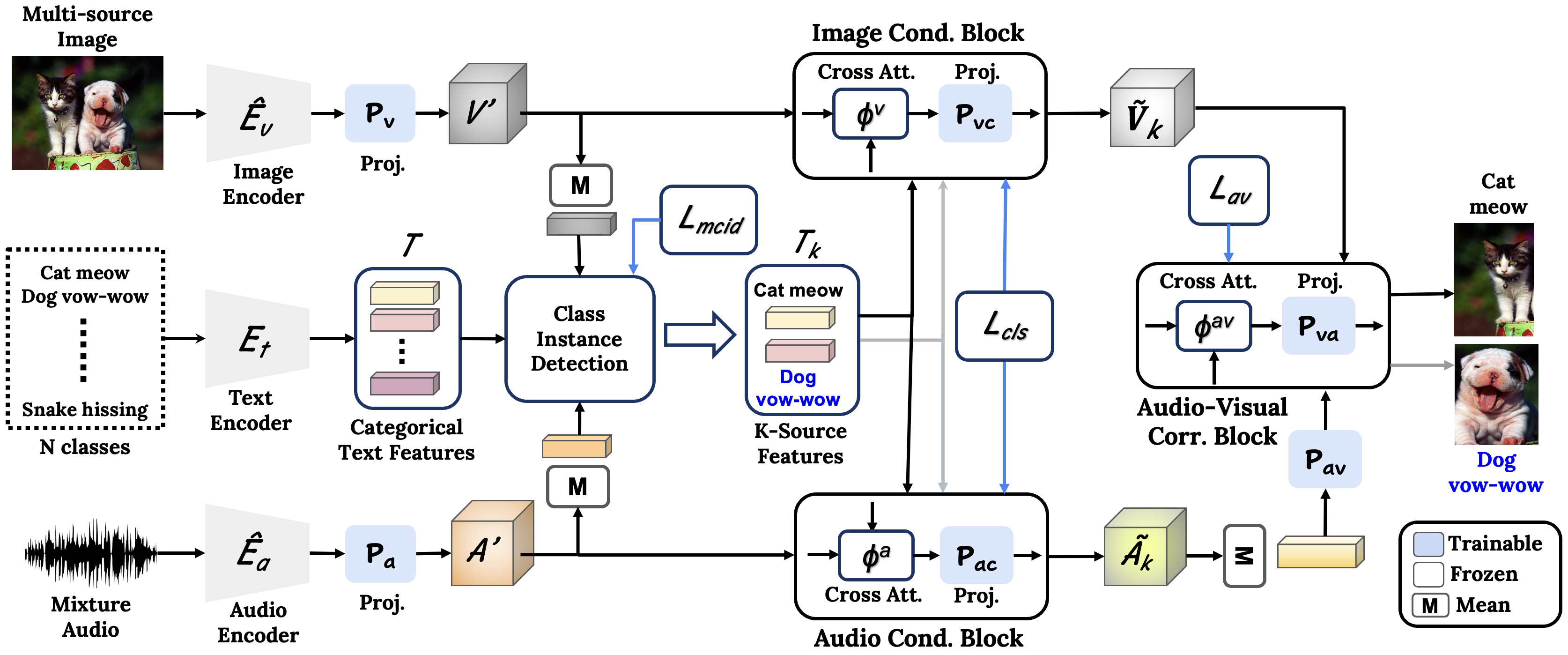}
    \vspace{-0.5em}
    \caption{The proposed text-guided visual sound source localization (T-VSL) framework (for $K = 2$).
    We use the text modality to disentangle the fine-grained audio-visual correspondence from mixtures.
    Initially, we detect the audio-visual class instances from multi-source mixtures using AudioCLIP joint embedding space.
    Later, categorical text features of each detected $K$ classes are used as coarse guidance in conditioning blocks to extract categorical visual and  audio features.
    Afterwards, cross-modal feature alignment on extracted categorical features is performed in audio-visual correspondence block.
    Finally, cosine similarity of mean categorical audio features, and aligned visual features are used recursively to extract localization map of each class.
    }
    \label{fig: main_img}
    \vspace{-1em}
\end{figure*}

\noindent\textbf{Visual Sound Source Localization.}
Visual sound source localization aims to locate the visual regions of sounding objects in video frames.
Prior work explored diverse methods for audio-visual feature alignment, including traditional statistical machine learning approaches~\cite{hershey1999audio,fisher2000learning,kidron2005pixels} to latest deep neural nets~\cite{Senocak2018learning,hu2019deep,Afouras2020selfsupervised,qian2020multiple,chen2021localizing,arda2022learning,mo2022EZVSL,mo2022SLAVC}.
However, most of these self-supervised and weakly-supervised methods are mostly designed for localizing single-source sounds.
A two-stream architecture is proposed in Attention10k~\cite{Senocak2018learning} which incorporates cross-attention to locate sounding sources.
Later, motion cues present in videos are incorporated for finer localization~\cite{Afouras2020selfsupervised}.
Recently, various contrastive learning based methods are explored in LVS~\cite{chen2021localizing}, EZ-VSL~\cite{mo2022EZVSL}, and SLAVC~\cite{mo2022SLAVC} for single-source localization. Despite their promising performance, these single-source localization methods can hardly discriminate audio-visual correspondence in multi-source mixtures. 

Sounds mostly appear in mixtures in the presence of significant background noises.
The main challenges of multi-source localization is to disentangle the audio-visual correspondence in mixtures without having access to clean single-source audio-visual pairs.
Several recent works have explored multi-source localization with explicit approaches for mixtures~\cite{qian2020multiple,hu2020dsol,hu2022mix, mo2023audio}.
To selectively isolate the silent objects in visual frames, DSOL~\cite{hu2020dsol} proposed a two-stage weakly-supervised training method.
Mix-and-localize~\cite{hu2022mix} proposed a self-supervised training with contrastive random walk to inherently align each sounding source with its visual representation.
Recently, AVGN~\cite{mo2023audio} introduced weakly-supervised grouping of category-aware features with learnable prompts.
However, these approaches tackle the audio-visual correspondence in mixtures without any explicit guidance for single-source thereby resulting in sub-optimal performance.
In contrast, we leverage text representation of fine-grained sounding sources to disentangle one-to-one audio-visual correspondence from multi-source mixtures.
To the best of our knowledge, this work is the first to utilize text representation of sound sources for solving multi-source localization.

\noindent\textbf{Audio, Visual, and Text Grounding.}
Large-scale pre-training on web-scraped data has been used to generate grounded features across modalities.
CLIP~\cite{radford2021learning} first introduced vision-language grounding using $400$M image-text pairs, which has been extensively studied for zero-shot classification~\cite{goyal2023finetune, wang2023clipn, guo2023calip, ali2023clip, jeong2023winclip, jeong2023winclip, esmaeilpour2022zero}, text-video retrieval~\cite{luo2022clip4clip, deng2023prompt, alpay2023multimodal}, open-vocabulary segmentation~\cite{luo2023segclip, xu2023learning, qin2023freeseg, ding2022open, ghiasi2022scaling, zhang2023simple, mukhoti2023open} and object detection~\cite{bravo2022localized, wang2023learning, minderer2022simple, du2022learning}.
Recently, AudioCLIP~\cite{guzhov2022audioclip} and Wav2CLIP~\cite{wu2022wav2clip} introduced additional audio grounding in original CLIP model by learning on image, text, and audio triplets.
Very recently, a CLIP-based single source localization method~\cite{park2023can} is introduced that attempts replacing the text-conditioning branch of a supervised image segmentation baseline, trained with dense supervision, with audio conditioning. 
In relation with this concurrent work, our focus lies primarily on weakly supervised sound source localization in multi-source mixtures, leveraging the self-supervised AudioCLIP model.

\section{Method}

Given a mixture of audio and a video frame, our objective is to spatially localize each sounding object in the frame. 
We propose a novel text-guided multi-source localization framework to disentangle the fine-grained audio-visual  correspondence, which is illustrated in Fig.~\ref{fig: main_img}. 
To suppress the background noises, we initially detect the common class instances in both audio and visual modality  (Sec.~\ref{sec:avid}).
In particular, we leverage tri-modal AudioCLIP encoders to detect $K$ visual-sounding class instances from $N$ classes ($K \leq N$).
The primary challenge in multi-source localization lies in establishing one-to-one audio-visual correspondences for individual sources without access to single-source samples.
To overcome this, we leverage the text modality to generate coarse representation of the $K$ detected sources in the mixture.
Later, by exploiting the multi-modal grounding of AudioCLIP, we extract categorical audio and visual features from multi-source representation using text guidance (Sec.~\ref{sec:avcn}).
Rather than aligning audio-visual mixture features as prior work, we introduce the categorical audio-visual feature alignment to further enhance one-to-one correspondence (Sec.~\ref{sec:avcrn}). 
Finally, cosine similarity of categorical audio features and aligned visual features is used to extract semantic localization maps.

\subsection{Preliminaries}
In this section, we formulate the multi-source localization problem, and revisit the AudioCLIP~\cite{guzhov2022audioclip} for audio, visual, and text grounding.
\vspace{2mm}

\noindent\textbf{Problem Formulation.}
Let's assume the dataset contains a total of $N$ classes of sounding events across all videos.
Given a video frame and an audio mixture, our objective is to localize $K (K \leq N)$ sound sources within the frame. 
For each audio-visual mixture pair, we can extract $Y = \{y_i \}_{i=1}^{N}$ binary ground truth labels, along with categorical text representations for the on-screen sound sources in the mixture.
During training, we are limited to video-level sounding class entities, excluding bounding boxes or mask-level annotations, thus constituting weakly-supervised learning.

\vspace{2mm}
\noindent\textbf{Revisit AudioCLIP for multi-modal grounding.}
AudioCLIP~\cite{guzhov2022audioclip} learns a joint embedding space of audio, visual, and text triplets through contrastive learning. 
Let's consider the dataset $\mathcal{D} = \{a, v, t\}_{i=1}^{Z}$ containing audio spectrogram $a \in \mathbb{R}^{M \times F}$, text $t \in \mathbb{R}^{l}$, and image $v \in \mathbb{R}^{C \times H \times W}$ triplets.
Also, separate audio encoder ($E_a: \mathbb{R}^{M\times F} \rightarrow \mathbb{R}^{D}$), visual encoder ($E_v: \mathbb{R}^{C\times H\times W} \rightarrow \mathbb{R}^{D}$), and text encoder ($E_t: \mathbb{R}^{l} \rightarrow \mathbb{R}^{D}$) are used for each modality. Hence, the extracted audio, visual, and text features are represented by $A =  E_a(a)$, $V =  E_v(v)$, and $T =  E_t(t)$, respectively, where $A, V, T \in  \mathbb{R}^{D}$. Later, InfoNCE contrastive loss ($\mathcal{L}_{\mbox{cnt}}$)~\cite{oord2018representation} across each modality pair is used for feature grounding given by:
\begin{equation}
\begin{aligned}\label{eq:ac_loss}
    & \mathcal{L}_{\mbox{aclip}} = \mathcal{L}_{\mbox{cnt}} (A, T) + \mathcal{L}_{\mbox{cnt}} (V, T) + \mathcal{L}_{\mbox{cnt}} (A, V)
\end{aligned}
\end{equation}
This training objective, combined with large-scale pre-training on web-scraped data, establishes a grounded joint embedding space across all three modalities.

\subsection{Audio-Visual Class Instance Detection}\label{sec:avid}

The target sounding objects for localization must be present in both the audio mixture and the corresponding video frame. However, audio mixtures may contain noise from off-screen background sources, while video frames may include non-sounding objects. To address these challenges, we first detect the class instances of visible sounding objects in the frame. Then, we utilize this detection to suppress redundant features in the mixture, ensuring that the localization focuses on the relevant sounding objects.

\noindent\textbf{Token extraction from Uni-modal encoders.} 
Multiple sounding objects can be spatially located in different regions of the reference frame, while the audio mixture may contain the temporally distributed signal of different sources. 
This makes it challenging to detect multiple class instances with uni-dimensional single-token features generated from AudioCLIP encoders. 
Instead of a single pooled token representation, we extract numerous patch tokens from audio and visual encoders with simple modifications, such that  $\widehat{E}_a: \mathbb{R}^{M\times F} \rightarrow \mathbb{R}^{n_a \times D}$ and $\widehat{E}_v: \mathbb{R}^{C\times H\times W} \rightarrow \mathbb{R}^{n_v \times D}$. 
Here, $n_a = m \times f$ and $n_v = h \times w$ represent the number of audio and visual patch tokens, respectively. 
Afterwards, we apply linear projectors on these extracted patch tokens, such that $P_a: \mathbb{R}^{n_a \times D} \rightarrow \mathbb{R}^{n_a \times D}$ and $P_v: \mathbb{R}^{n_v \times D} \rightarrow \mathbb{R}^{n_v \times D}$.
Thus, extracted audio and visual patch tokens are represented as ${A'} = P_a(\widehat{E}_a(a))$, ${V'} = P_v(\widehat{E}_v(v))$, respectively, where ${A'} \in \mathbb{R}^{n_a\times D}$ and ${V'} \in \mathbb{R}^{n_v\times D}$.
Simultaneously, categorical text features $\mathcal{T} = {\{e^t_i\}_{i=1}^{N}} \in \mathbb{R}^{N \times D}$ representing all $N$ sound source classes in the dataset are extracted. 

\noindent\textbf{Multi-label class instance detection.} 
With the categorical text features  $\mathcal{T}$, mixture audio ${A'}$, and visual ${V'}$ patch tokens, we detect class entities present in the mixture. Initially, we extract the mean-pooled audio-visual features $X' \in \mathbb{R}^{2D}$ by simple concatenation, and then, apply fusion projector $P_f : \mathbb{R}^{2D} \rightarrow \mathbb{R}^{D}$ to extract mixture audio-visual features $F_{av} \in \mathbb{R}^{D}$. Finally, cosine similarity between $N$ class text features $\mathcal{T}$ and $F_{av}$ is used to detect mixture class entities $\widehat{Y} \in \mathbb{R}^{N}$, given by:
\begin{equation}
\begin{aligned}
    & {F}_{av} = P_f (X'), \ {X}' = [\mathtt{Mean(}A') \oplus \mathtt{Mean}(V')],\\
    & \widehat{Y} =  \mathtt{sim} \left(\mathcal{T}, {F_{av}}\right)
\end{aligned}
\end{equation}
where $[\ \oplus \ ]$ denotes the concatenation operator, and $\mathtt{sim}(X, Y)$ denotes cosine similarity between $X$ and $Y$.  We formulate a multi-label classification objective $\mathcal{L}_{\mbox{mcid}}$ by applying binary cross-entropy loss $\mathcal{L}_{\mbox{bce}}(\cdot)$ on each prediction using video-level ground-truth labels $Y \in \mathbb{R}^{N}$ as:
\begin{equation}
    \mathcal{L}_{\mbox{mcid}} = \sum_{i=1}^N \mathcal{L}_{\mbox{bce}} (y_i, \widehat{y}_i)
\end{equation}
In particular, we train audio, visual, and fusion projectors ($P_a, P_v, P_f$) on noisy multi-source data keeping the AudioCLIP backbones frozen. Hence, these projectors assist in extracting relevant foreground features from noisy audio and visual mixture features extracted from AudioCLIP.

\begin{table*}[t]
	\renewcommand\tabcolsep{6.0pt}
	\centering
	\scalebox{0.8}{
		\begin{tabular}{l|ccc|ccc|cccc}
			\toprule
			\multirow{2}{*}{Method} & \multicolumn{3}{c|}{VGGSound-Single} & \multicolumn{3}{c|}{VGGSound-Instruments} & \multicolumn{3}{c}{MUSIC-Solo} \\
			&  AP(\%) & IoU@0.5(\%) & AUC(\%) &  AP(\%) & IoU@0.3(\%) & AUC(\%) & AP(\%) & IoU@0.5(\%) & AUC(\%) \\ 	
			\midrule
                   \textit{Current SOTA Based Methods} \\

                  OTS~\cite{Arandjelovic2018ots} \scriptsize ECCV18  & 38.9 & 43.2 & 43.7 & 50.9 & 56.9 & 38.4 & 76.2 & 57.8 & 48.9 \\
                CoarsetoFine~\cite{qian2020multiple} \scriptsize ECCV20 & 39.4 & 43.1 & 44.8 & 51.1 & 57.5 & 39.1 & 76.3 & 58.1 & 49.7 \\
              
			LVS~\cite{chen2021localizing} \scriptsize CVPR21 & 39.7 & 44.6 & 45.3 & 51.5 & 57.9 & 39.4 & 76.1 & 57.1 & 49.3 \\

    		EZ-VSL~\cite{mo2022EZVSL} \scriptsize ECCV22  & 40.6 & 45.1 & 47.2 & 52.3 & 58.8 & 39.9 & 78.4 & 58.7 & 51.1 \\

                Mix-and-Localize~\cite{hu2022mix} \scriptsize CVPR22  & 41.8 & 45.6 & 47.4 & 53.8 & 59.3 & 41.5 & 78.9 & 59.9 & 52.7 \\

                DSOL~\cite{hu2020dsol} \scriptsize NeurIPS20 & -- & 46.8 & 47.9 & -- & 61.6 & 43.7 & --   & 62.7 & 54.6 \\

                MarginNCE~\cite{park2023marginnce} \scriptsize ICASSP23  & 42.5 & 46.9 & 48.3 & 56.9 & 61.8 & 44.2 & 83.1 & 63.3 & 55.2 \\
                FNAC~\cite{fnac} \scriptsize CVPR23  & 43.3 & 48.4 & 49.1 & 57.2 & 63.2 & 45.3 & 84.6 & 64.5 & 56.4 \\
                AVGN~\cite{mo2023audio} \scriptsize CVPR23  & 44.1 & 49.6 & 49.5 & 59.3 & 64.7 & 46.1 & 85.4 & 65.8 & 56.9 \\
                Alignment~\cite{alignment} \scriptsize ICCV23  & 45.3 & 50.8 & 50.2 & 60.4 & 65.8 & 48.7 & 86.1 & 66.4 & 57.2 \\
                \midrule
                \textit{CLIP-Based Baseline Methods} \\
                AudioCLIP~\cite{guzhov2022audioclip} \scriptsize ICASSP22 & 42.8 & 47.4 & 48.5 & 58.3 & 62.7 & 45.2 & 83.8 & 63.1 & 55.7 \\

                Wav2CLIP ~\cite{wu2022wav2clip} \scriptsize ICASSP22 & 39.3 & 43.6 & 44.7 & 53.8 & 57.2 & 42.0 & 78.9 & 59.9 & 51.2 \\
                \midrule
                T-VSL (Ours) & \textbf{48.1} & \textbf{53.7} & \textbf{52.9} & \textbf{64.6} & \textbf{69.5} & \textbf{51.4} & \textbf{88.2} & \textbf{68.5} & \textbf{60.1} \\
			\bottomrule
			\end{tabular}}
    \caption{Performance comparison of single-source localization on VGGSound-single, VGGSound-Instruments, and MUSIC-Solo datasets. For the fair comparison, we use pre-trained AudioCLIP audio and image encoders for baseline methods.}
   \label{tab: exp_sota_single}
  \vspace{-1em}
\end{table*}

\subsection{Audio and Visual Conditioning Blocks}\label{sec:avcn}

Establishing one-to-one correspondences for each source in multi-source mixtures is particularly challenging, as both audio and visual features contain mixed representations of various sources.
Prior works~\cite{mo2023audio, hu2022mix} on multi-source localization have attempted to learn audio-visual alignment intrinsically without explicit single-source guidance. 
However, in the absence of single-source data, these methods struggle to adequately disentangle multi-source features.
In contrast, we leverage the categorical text features of $K$ classes present in the mixture as coarse guidance for fine-grained feature disentanglement from multi-source mixtures.
We use ground truth weak labels $Y$ of representative audios in the mixture to extract text embedding of $K$ visual sounding sources $\mathcal{T}_K = \{e^t_k\}_{k=1}^K \subseteq \mathcal{T}$ where $e^t_k \in \mathbb{R}^{1 \times D}$. 

With the categorical text features $\mathcal{T}_K$ of $K$ sources, we independently disentangle the multi-object visual patch tokens ${V'}$ and mixture audio patch tokens ${A'}$ by using cross-attention $\phi^v (\cdot)$ and $\phi^a (\cdot)$, respectively, as:
\begin{equation}
\begin{aligned}
    & \widetilde{V}_k = \phi^v(V', e^t_k, V'), \ \mathnormal{f}^v_k = \mathtt{Mean}(\widetilde{V}_k),  \\
    & \widetilde{\mathnormal{f}}^v_k = P_{vc}(\mathnormal{f}^v_k), \ \ \forall k = \{1, \dots, K\}
\end{aligned}
\end{equation}
\begin{equation}
\begin{aligned}
    & \widetilde{A}_k = \phi^a(A', e^t_k, A'), \ \mathnormal{f}^a_k = \mathtt{Mean}(\widetilde{A}_k),  \\
    & \widetilde{\mathnormal{f}}^a_k = P_{ac}(\mathnormal{f}^a_k), \ \ \forall k = \{1, \dots, K\}
\end{aligned}
\end{equation}
\begin{equation}
\begin{aligned}
     \phi(A, B, C) = \left(\dfrac{A B^\top}{\norm{A}\norm{B}}\right) \odot C 
\end{aligned}
\end{equation}
where  $\widetilde{\mathnormal{f}}^a_k, \widetilde{\mathnormal{f}}^v_k, \mathnormal{f}^a, \mathnormal{f}^v \in\mathbb{R}^{1\times D}$, $\widetilde{V}_k \in \mathbb{R}^{n_v\times D}$,  $\widetilde{A}_k \in \mathbb{R}^{n_a\times D}$, and $P_{ac}, P_{vc} : \mathbb{R}^D \rightarrow \mathbb{R}^D$ are linear audio and visual projectors on conditioned mean features, respectively, and $\odot$ is the Hadamard product. 

Additionally, to guide extracting categorical audio ($\widetilde{A}_k$) and visual ($\widetilde{V}_k$) patch tokens, we introduce coarse supervision on the output of projected mean patch tokens $\widetilde{\mathnormal{f}}^a_k$, $\widetilde{\mathnormal{f}}^v_k$ using $N$-class text embedding $\mathcal{T} \in \mathbb{R}^{N \times D}$. Hence, the class conditioning loss ($\mathcal{L}_{\mbox{cls}}$) is given by:
\begin{equation}
    \begin{aligned}
        & \mathbf{e}^v_k =  \mathtt{sim}(\mathcal{T}, \widetilde{\mathnormal{f}}^v_k);\ \mathbf{e}^a_k =  \mathtt{sim}(\mathcal{T}, \widetilde{\mathnormal{f}}^a_k) \\
        & \mathcal{L}_{\mbox{cls}} = \sum_{k=1}^K \mathcal{L}_{\mbox{ce}}( \mathbf{e}^v_k, \mathbf{h}_k) + \mathcal{L}_{\mbox{ce}}(\mathbf{e}^a_k, \mathbf{h}_k)
    \end{aligned}
\end{equation}
where $\mathbf{h}_k \in \mathbb{R}^{N}$ denotes an one-hot encoding of the class label, and $\mathbf{e}^v_k \in \mathbb{R}^{N}$, $\mathbf{e}^a_k \in \mathbb{R}^{N}$ are visual and audio class predictions for each $k^{th}$ source, respectively.
 
\subsection{Audio-Visual Correspondence Block}\label{sec:avcrn}
While text representation provides coarse guidance for disentangling fine-grained spatio-temporal audio and visual features, it has inherent limitations in practice.
For example, multiple instances of a sounding object class can appear in the visual frame, including silent instances, which are difficult to distinguish using simple text class representation. 
To address this challenge, we introduce an audio-visual correspondence block to further refine the alignment of audio and visual features based on extracted categorical features. 

With the disentangled categorical audio ($\widetilde{A}_k$) and visual ($\widetilde{V}_k$) patch tokens, we apply cross-attention $\phi^{av}(\cdot)$ to enhance audio-visual patch alignment as:
\begin{equation}
\begin{aligned}
    & g_k^a = \mathtt{Mean}(\widetilde{A}_k), \ \widehat{\mathnormal{g}}^a_k = P_{av}(\mathnormal{g}^a_k) \\
    & \widehat{V}_k = \phi^{av}(\widetilde{V}_k, \widehat{\mathnormal{g}}^a_k, \widetilde{V}_k), \ \mathnormal{g}^v_k = \mathtt{Mean}(\widehat{V}_k) \\
    & \widehat{\mathnormal{g}}^v_k = P_{va}(\mathnormal{g}^v_k), \ \forall k = \{1, \dots, K\}
\end{aligned}
\end{equation}
where $P_{av}, P_{va} : \mathbb{R}^D \rightarrow \mathbb{R}^D$ are linear projectors.
Afterwards, to correspond between projected mean categorical audio and visual features $\widehat{\mathnormal{g}}^a_k, \widehat{\mathnormal{g}}^v_k \in \mathbb{R}^{D}$, we apply audio-visual contrastive loss ($\mathcal{L}_{\mbox{av}}$) given by:
\begin{equation}
\begin{aligned}
    & \mathcal{L}_{\mbox{av}} = \mathcal{L}_{\mbox{cnt}}(\widehat{g}_k^v, \widehat{g}_k^a) \\
        & \mathcal{L}_{\mbox{total}} = \mathcal{L}_{\mbox{av}} + \mathcal{L}_{\mbox{cls}} + \mathcal{L}_{\mbox{mcid}}
\end{aligned}    
\end{equation}
where $\mathcal{L}_{\mbox{total}}$ is the total accumulated loss.
\vspace{2mm}

\noindent \textbf{Inference:}
During inference, we estimate cosine similarity across projected mean audio feature $\widehat{g}_k^a \in \mathbb{R}^{D}$ and aligned visual patch tokens ($\widehat{V}_k = \{\widehat{v}_k^j\}_{j=1}^{n_v}, \widehat{v}_k^j \in \mathbb{R}^{D}$).
Finally, after reshaping and up-scaling through bilinear interpolation, we generate $K$-source heat-maps $\mathbf{{H}} \in \mathbb{R}^{K \times H \times W}$.

\begin{table*}[t]
	\renewcommand\tabcolsep{4.0pt}
	\centering
	\scalebox{0.78}{
		\begin{tabular}{l|ccc|ccc|ccc}
			\toprule
			\multirow{2}{*}{Method} &  \multicolumn{3}{c|}{VGGSound-Duet} & \multicolumn{3}{c|}{VGGSound-Instruments} & \multicolumn{3}{c}{MUSIC-Duet} \\
			& CAP(\%) & CIoU@0.3(\%) & AUC(\%) & CAP(\%) & CIoU@0.1(\%) & AUC(\%) & CAP(\%) & CIoU@0.3(\%) & AUC(\%) \\ 	
			\midrule
                \textit{Current SOTA Single-Source Methods} \\
                OTS~\cite{Arandjelovic2018ots} \scriptsize ECCV18 & 23.9 & 26.4 & 27.5 & 28.7 & 77.8 & 18.2 & 51.2 & 29.4 & 24.3 \\
                CoarsetoFine~\cite{qian2020multiple} \scriptsize ECCV20 & --  & 28.4 & 27.9 & --  & 78.8 & 19.7 & --& 31.8 & 24.1 \\
			LVS~\cite{chen2021localizing} \scriptsize CVPR21 & -- & 31.8 & 30.1 & -- & 81.1 & 23.2 & --& 33.1 & 26.9 \\
			EZ-VSL~\cite{mo2022EZVSL} \scriptsize ECCV22 & -- & 32.4 & 30.6 & -- & 80.3 & 22.9 & --& 34.3 & 26.7  \\
                MarginNCE~\cite{park2023marginnce} \scriptsize ICASSP23 & {27.1} & {33.2} & {31.2} & {33.5} & {82.3} & {24.7} & {55.1}	& {36.1} & {28.5} \\
                FNAC~\cite{fnac} \scriptsize CVPR23 & {29.8} & {35.1} & {33.4} & {35.4} & {84.1} & {26.8} & {57.3}	& {37.9} & {30.6} \\
                Alignment~\cite{alignment} \scriptsize ICCV23 & {30.4} & {35.8} & {33.9} & {35.6} & {84.9} & {27.1} & {57.8}	& {38.3} & {30.9} \\
                \midrule
                \textit{Current SOTA Multi-Source Methods} \\
                Mix-and-Localize~\cite{hu2022mix} \scriptsize CVPR22 & 29.5 & 35.6 & 34.1 & 36.3 & 84.5 & 26.7 & 57.4 & 38.1 & 30.7 \\
                DSOL~\cite{hu2020dsol} \scriptsize NeurIPS20 & -- & 36.9 & 34.6 & -- & 85.9 & 28.4 & -- & 38.8 & 31.3 \\
                AVGN~\cite{mo2023audio} \scriptsize CVPR23 & {31.9} & {37.8} & {35.4} & {38.1} & {86.7} & {28.8} & {59.2}	& {39.3} & {32.4} \\
                \midrule
                \textit{CLIP-Based Baseline Methods} \\
                AudioCLIP~\cite{guzhov2022audioclip} \scriptsize ICASSP22 & 28.4 & 34.9 & 32.8 & 34.7 & {83.8} & {25.9} & {56.1} & {36.9} & {29.2} \\   
                Wav2CLIP~\cite{wu2022wav2clip} \scriptsize ICASSP22 & 26.3 & 31.4 & 28.9 & 31.2 & {80.3} & {22.2} & {52.6} & {32.2} & {27.2} \\        
                 \midrule
                T-VSL (ours) & \textbf{35.7} & \textbf{40.1} & \textbf{37.9} & \textbf{41.8} & \textbf{89.6} & \textbf{31.5} & \textbf{62.9}	& \textbf{43.2} & \textbf{35.9} \\
			\bottomrule
			\end{tabular}}
   \vspace{-0.5em}
   \caption{Performance comparison of multi-source localization on VGGSound-Duet, VGGSound-Instruments, and MUSIC-Duet datasets. For the fair comparison, we use pre-trained AudioCLIP audio and image encoders for baseline methods.}
   \label{tab: exp_sota_multi}
   \vspace{-1em}
\end{table*}

\vspace{-0.5em}
\section{Results and Discussions}
\vspace{-0.5em}

\subsection{Experimental setup}
\label{datasets}

\noindent\textbf{Datasets.}
We used MUSIC, VGGSound-Instruments, and VGGSound datasets for the performance evaluation\footnote{Since many videos are no longer available for public use, these datasets become smaller. We use the same data split for all baselines and reproduce them under the same setting for a fair comparison.}.
MUSIC~\cite{zhao2018the} dataset contains $445$ solo music videos of $11$ instruments and $142$ duet music videos of $8$ instruments from YouTube. Following prior works~\cite{mo2023audio}, we use MUSIC-Solo for single-source and MUSIC-Duet for multi-source localization. From MUSIC-Solo, we use $350$ videos for training and remaining $95$ for evaluation. For MUSIC-Duet, we use $120$ videos for training and remaining $22$ videos for evaluation.
VGGSound-Instruments~\cite{hu2022mix} is a subset of VGGSound dataset~\cite{chen2020vggsound} containing around $32$k video clips from $37$ music instruments of $10$s duration.
Apart from the musical instrument datasets, we gather around $150$k videos of  $10$s duration from original VGGSound dataset~\cite{chen2020vggsound} representing single-source sounds of $221$ categories, such as animals, people, nature, instruments, etc.
For the single-source evaluation on VGGSound, we use the full test set of $5158$ videos, while on VGGSound-Instruments, we use $446$ videos following prior work~\cite{mo2023audio, hu2022mix}.
For the multi-source training and evaluation in both VGGSound and VGGSound-Instruments, we randomly concatenate two frames from different sounding sources to produce the multi-source input image with a size of $448\times224$, and mix the single-source audios, following prior work~\cite{mo2023audio, hu2022mix}.

\noindent\textbf{Evaluation Metrics.}
For evaluating single-source localization performance,  we use Intersection over Union (IoU), average precision (AP), and Area Under Curve (AUC) following prior work~\cite{mo2023audio, hu2022mix}.
For the multi-source localization evaluation, we use Class-aware Average precision (CAP),
Class-aware IoU (CIoU), and Area Under Curve (AUC)  following prior work~\cite{mo2023audio, hu2022mix}.
Similarly, for fair comparison with the prior work~\cite{mo2023audio}, we use the same thresholds in these metrics: we use IoU@0.5 and CIoU@0.3 for the MUSIC-Solo and MUSIC-Duet datasets, IoU@0.3 and IoU@0.5 for single-source VGGSound-Instruments and VGGSound datasets, and CIoU@0.1 and CIoU@0.3 for multi-source VGGSound-Instruments and VGGSound.

\noindent\textbf{Implementation details.}
We use the AudioCLIP~\cite{guzhov2022audioclip} pre-trained encoders containing ResNet-50~\cite{hu2019deep} image encoder, ESResNeXt~\cite{guzhov2021esresne} audio encoder, and transformer~\cite{radford2019language} based text encoder.
For the fair comparison, we reproduce the result of existing methods with AudioCLIP image and audio encoders, instead of using separately pre-trained encoders.
We use $224\times 224$ resolution for the single-source input image,
$D=1024$, $n_v = 49$ for $7\times 7 (h\times w)$ spatial maps generated from the visual encoder, and $n_a=60$ for $10 \times 6 (m\times f)$ time-frequency map generated from the audio encoder. We use Adam optimizer~\cite{kingma2014adam} with a batch size of $256$ and with a learning rate of $1e-4$.

\noindent\textbf{CLIP-Based Baseline methods.}
We adopt several CLIP-based baseline methods for the sound source localization task, which are described as follows:
\begin{itemize}
    \item \textbf{AudioCLIP}~\cite{guzhov2022audioclip}: We fine-tune pre-trained image and audio encoders of AudioCLIP on the target dataset. We estimate the localization heatmap for the sounding sources using the similar cosine similarity across patch tokens. 
    \item \textbf{Wav2CLIP}~\cite{wu2022wav2clip}: Similar to AudioCLIP baseline, we experiment with Wav2CLIP audio and image encoders to extract localization heatmaps of sounding sources.
\end{itemize}

\begin{figure*}[t]
\centering
\includegraphics[width=0.9\linewidth]{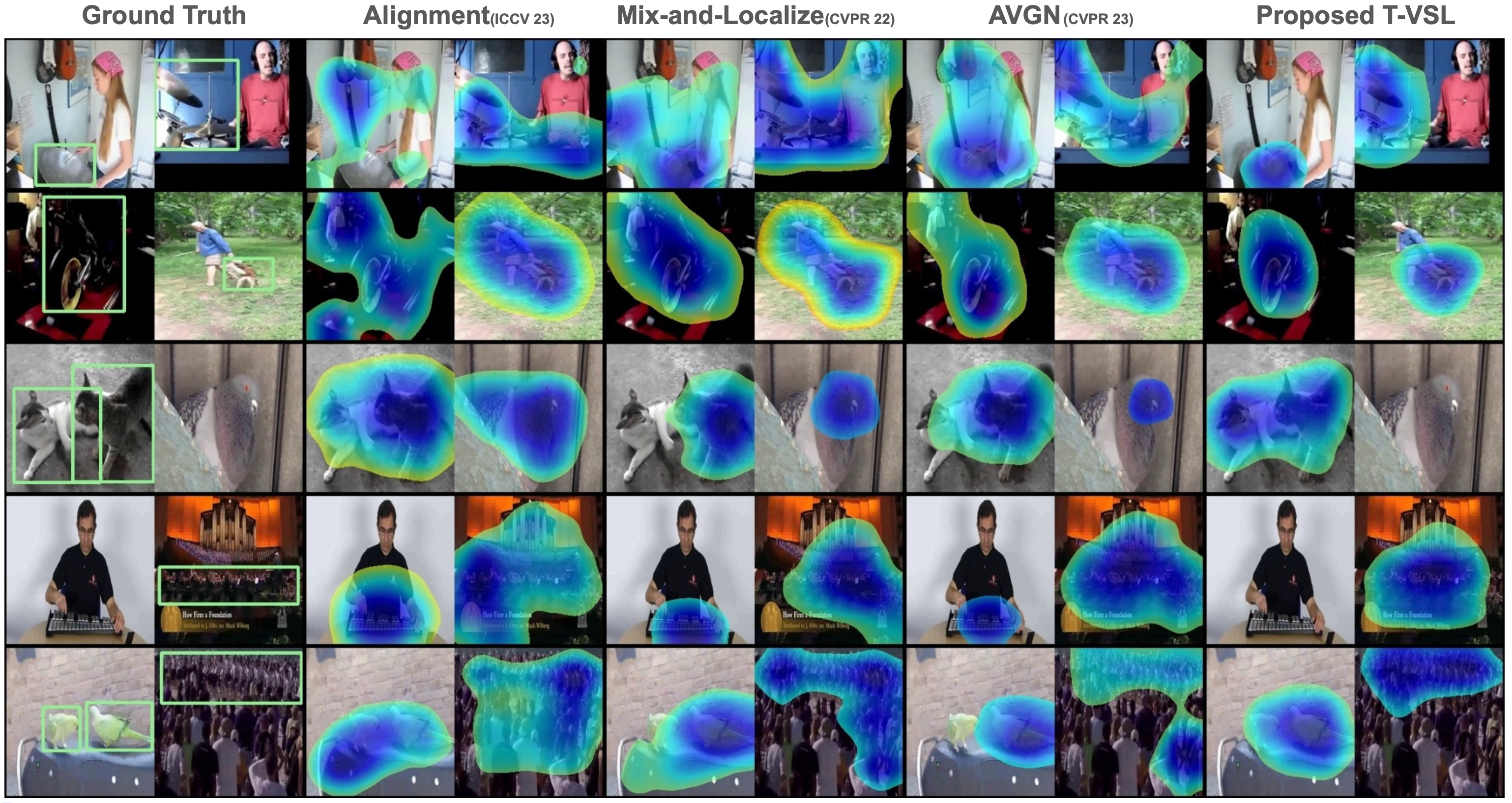}
\caption{
We present qualitative comparisons on challenging multi-source localization with SOTA single and multi-source baseline methods. Here, \textcolor{blue}{blue} color represents high-attention values to the sounding object, and \textcolor{red}{red} color represents low-attention values. The proposed T-VSL can selectively isolate the sounding regions from the background and generates more precise localization maps for sounding sources.}
\label{fig: vis_source}
\vspace{-1em}
\end{figure*}

\subsection{Comparison to SOTA Baselines}\label{sec:exp}
\noindent \textbf{Single-source localization:}
We present the quantitative results on single-source localization in Tab.~\ref{tab: exp_sota_single}.
We apply the AudioCLIP image and audio encoders in all baseline methods for the fair comparison with our methods.
We observe some performance improvements with several SOTA methods~\cite{alignment, mo2023audio, fnac} over the AudioCLIP baseline.
However, we also notice some reduction in performance from the AudioCLIP baseline in several methods~\cite{Arandjelovic2018ots, mo2022EZVSL, hu2022mix}. We hypothesize that these can be due to the alterations of AudioCLIP grounding with the modified contrastive loss.
In contrast, by utilizing all three modalities for sound source localization, the proposed T-VSL achieves the best performance over all CLIP-based baselines as well as SOTA methods.
T-VSL outperforms the current SOTA single source localization method Alignment~\cite{alignment} by $+2.9$, $+3.7$, $+2.1$ higher IoUs on VGGSound-Single, VGGSound-Instruments, and MUSIC-solo datasets, respectively.
Moreover, T-VSL achieves $+6.3$, $+6.8$, and $+5.4$ higher IoUs over the AudioCLIP baseline~\cite{pacl}. 
We hypothesize that the enhanced refinement of noisy audio and visual features with text guidance in T-VSL effectively contributes to performance improvement.

\noindent \textbf{Multi-source localization:}
We present performance comparison in multi-source sound localization in Tab.~\ref{tab: exp_sota_multi}.
As before, we apply the AudioCLIP image and audio encoders in all baseline methods for the fair comparison with our methods.
We observe some performance improvements, mostly in SOTA multi-source methods~\cite{hu2022mix, mo2023audio, hu2020dsol}, over the AudioCLIP baseline. In most single source methods, we observe performance loss in multi-source cases over the baseline.
The proposed T-VSL achieves the best performance that outperforms the SOTA multi-source baseline AVGN~\cite{mo2023audio} by $+2.3$, $+2.9$, and $+3.9$ CIoUs on multi-source VGGSound-Duet, VGGSound-Instruments, and MUSIC-Duet datasets, respectively.
Also, T-VSL achieves $+5.2$, $+5.8$, and $+6.3$ higher CIoUs over the AudioCLIP baseline. 
These significant performance gains on multi-source benchmarks demonstrate the effectiveness of the proposed T-VSL in disentangling multi-source mixtures.

\noindent \textbf{Qualitative comparisons:}
In addition, we present qualitative comparisons on multi-source localization in Fig.~\ref{fig: vis_source} among single-source baseline Alignment~\cite{alignment}, self-supervised multi-source baseline Mix-and-Localize~\cite{hu2022mix}, weakly-supervised multi-source baseline AVGN~\cite{mo2023audio}, and proposed T-VSL.
Single-source SOTA baseline Alignment~\cite{alignment} struggles in multi-source localization due to the lack of audio-visual correspondence, mostly in the presence of noisy audios and silent visual objects.
However, without any guidance on fine-grained source localization, multi-source baselines~\cite{mo2023audio, hu2022mix} underperform on localizing challenging multi-source mixtures.
Notably, our method generates superior localization maps, which shows the effectiveness of T-VSL in multi-source feature disentanglement.

\vspace{-0.5em}
\subsection{Ablation on T-VSL building blocks}
\vspace{-0.5em}
We present the ablation study of different building blocks of T-VSL in Tab.~\ref{tab: exp_ablation} under both single-source and multi-source settings. 
The vanilla baseline contains AudioCLIP image and audio encoders that directly operate on the input data. 
By integrating the audio-visual correspondence (AVC) block on extracted mixture patch tokens, we notice considerable performance increase in single-source (by $+1.1$ AP and $+0.7$ IoU@0.5) and multi-source localization (by $+1.3$ CAP and $+1.4$ CIoU@0.3). 
By integrating the audio and image conditioning blocks following audio-visual instance detection for separating fine-grained class features, we observe additional improvements of $+1.8$ AP and $+2.3$ AP in single-source, and $+3.1$ CAP and $+3.6$ AP  in multi-source localization, respectively.
Finally, the combination of all building blocks results in the best performance that improves the baseline by $+5.3$ AP, and $6.3$ IoU@0.5 in single-source and by $+7.3$ CAP and $+5.2$ CIoU@0.3 in multi-source localization. This further demonstrates the effective disentanglement of audio-visual features facilitated by intermediate text guidance.

\begin{table}[t]
	\renewcommand\tabcolsep{6.0pt}
    \renewcommand{\arraystretch}{1.1}
	\centering
	\scalebox{0.7}{
		\begin{tabular}{cccccccc}
			\toprule
			\multicolumn{1}{c}{\multirow{2}{*}{\begin{tabular}[c]{@{}c@{}}Audio\\ Cond.\end{tabular}}} & \multicolumn{1}{c}{\multirow{2}{*}{\begin{tabular}[c]{@{}c@{}}Image\\ Cond.\end{tabular}}} & \multirow{2}{*}{AVC} & \multicolumn{2}{c}{VGGSound-Single} & \multicolumn{2}{c}{VGGSound-Duet} \\
                \cmidrule(lr){4-5} \cmidrule(lr){6-7}
			& & & AP(\%) & IoU@0.5(\%) & CAP(\%) & CIoU@0.3(\%)  \\ 	
			\midrule
                \xmark & \xmark & \xmark & 42.8 & 47.4 & 28.4 &  34.9 \\
                \xmark & \xmark & \cmark & 43.9 & 48.1 & 29.7 &  36.3 \\
			\cmark & \xmark & \cmark & 45.7 & 50.9 & 32.8 &  38.1 \\
			\xmark & \cmark & \cmark & 46.2 & 51.4 & 33.3 &  38.6 \\
			\cmark & \cmark & \cmark & \textbf{48.1} & \textbf{53.7} & \textbf{35.7} &  \textbf{40.1} \\
			\bottomrule
			\end{tabular}}
   \caption{Ablation study on audio conditioning block, image-conditioning block, and audio-visual correspondence block (AVC) of T-VSL in single-source and multi-source localization tasks.}
	\label{tab: exp_ablation}
			\vspace{-0.5em}
\end{table}

\begin{table}[t]
	\renewcommand\tabcolsep{6.0pt}
    \renewcommand{\arraystretch}{1.1}
	\centering
	\scalebox{0.8}{
            \begin{tabular}{ccccc}
            \toprule
            \begin{tabular}[c]{@{}c@{}}Target\\ Dataset\end{tabular} & \begin{tabular}[c]{@{}c@{}}Proposed\\ T-VSL \end{tabular} & \begin{tabular}[c]{@{}c@{}}Mix-and-\\ Localize~\cite{hu2022mix}\end{tabular} & \begin{tabular}[c]{@{}c@{}}Alignment \\ \cite{alignment} \end{tabular} & \begin{tabular}[c]{@{}c@{}}FNAC \\ \cite{fnac} \end{tabular} \\
            \midrule
            MUSIC-Duet                                               & \textbf{55.8}  & 50.9 & 51.7 & 51.2                                                             \\
            MUSIC-Solo                                               & \textbf{80.5}  & 75.1 & 77.6 & 77.4                                                       \\
            VGGSound-Instr.                                          & \textbf{81.7} & 76.8 & 76.5 & 76.1 \\
            \bottomrule
            \end{tabular}}
   \caption{Ablation on zero-shot transfer across single and multi-source datasets with 50\%-50\% class split.
    We report IoU@0.5(\%) for MUSIC-Solo, CIoU@0.3(\%) for MUSIC-Duet, and CIoU@0.1(\%) for VGGSound-Instruments. We use same AudioCLIP encoders for all baselines.
    }
	\label{tab: exp_zero_shot_transfer}
			\vspace{-1.5em}
\end{table}
\vspace{-0.5em}
\subsection{Zero-shot transfer across datasets}
\vspace{-0.5em}
We can perform zero-shot transfer across datasets with proposed T-VSL by simply replacing the $N$-class text prompts in Fig.~\ref{fig: main_img}. 
Since other weakly-supervised methods cannot perform zero-shot transfer, we present comparisons with self-supervised Mix-and-Localize~\cite{hu2022mix}, Alignment~\cite{alignment}, and FNAC~\cite{fnac} methods in Tab.~\ref{tab: exp_zero_shot_transfer}. 
To ensure a fair comparison, we conducted uniform training with 50\%-50\% class split for all methods with same AudioCLIP encoders.
We note that our method achieves significantly better performance on zero-shot transfer to target datasets than these compared approaches.
In particular, we achieve $+4.9$, $+5.4$, $+4.9$ improvements on target datasets than the multi-source Mix-and-Localize~\cite{hu2022mix}.
Also, we achieve $+4.1$, $+2.9$, and $+5.2$ higher scores than the SOTA single source baseline, Alignment~\cite{alignment}.
This result suggests that our T-VSL has superior generalization capabilities for zero-shot localization tasks in unseen classes.

\vspace{-0.5em}
\subsection{Robustness to a higher number of sources}
\vspace{-0.5em}
The proposed T-VSL is not limited by the number of sources present in training mixtures unlike prior work~\cite{hu2022mix}. 
We present comparative analysis with SOTA multi-source AVGN~\cite{mo2023audio}, and self-supervised single source methods FNAC~\cite{fnac} and Alignment~\cite{alignment}  in Tab.~\ref{tab: exp_ablation_higher_src} on robustness to higher number of test sources in VGGSound dataset.
We train all methods on the VGGSound-Single dataset  with same encoders for a fair comparison.
We note that our method consistently maintains significantly higher CIoU score irrespective of test scenarios, and also achieves notably smaller relative performance drop for a large number of sources in mixtures.
These results further demonstrate the robustness of the proposed T-VSL in disentangling multi-source mixtures.

\begin{table}[t]
	\renewcommand\tabcolsep{6.0pt}
    \renewcommand{\arraystretch}{1.1}
	\centering
	\scalebox{0.9}{
            \begin{tabular}{ccccc}
            \toprule
            \begin{tabular}[c]{@{}c@{}} Test \\ Source No. \end{tabular} & \begin{tabular}[c]{@{}c@{}}Proposed\\ T-VSL \end{tabular} & \begin{tabular}[c]{@{}c@{}}AVGN\\~\cite{mo2023audio}\end{tabular} & \begin{tabular}[c]{@{}c@{}}Alignment\\~\cite{alignment}\end{tabular} & \begin{tabular}[c]{@{}c@{}}FNAC\\~\cite{fnac}\end{tabular}\\
            \midrule
            2                                               &  \textbf{39.6} & 36.8     &  35.4  & 34.1                                                     \\
            3                                          & \textbf{35.7}  & 30.8  & 28.9   & 27.1 \\
            4                                               & \textbf{29.5}  & 22.4    & 19.3    &   18.8                                                      \\
            \bottomrule
            \end{tabular}}
   \vspace{-0.5em}
   \caption{Robustness to higher number of sources in VGGSound dataset when trained with single-source data. CIoU@0.3(\%) score is reported. Same AudioCLIP encoders are used for all baselines.}
	\label{tab: exp_ablation_higher_src}
	\vspace{-0.5em}
\end{table}
\begin{table}[t]
	\renewcommand\tabcolsep{6.0pt}
    \renewcommand{\arraystretch}{1.1}
	\centering
	\scalebox{0.85}{
		\begin{tabular}{ccccc}
			\toprule
			\multicolumn{1}{c}{\multirow{2}{*}{\begin{tabular}[c]{@{}c@{}}Prompt\\ Length\end{tabular}}}  & \multicolumn{2}{c}{VGGSound-Single} & \multicolumn{2}{c}{VGGSound-Duet} \\
                \cmidrule(lr){2-3} \cmidrule(lr){4-5}
			& AP(\%) & IoU@0.5(\%) & CAP(\%) & CIoU@0.3(\%)  \\ 	
			\midrule

                0 & 48.1 & 53.7 & 35.7 &  40.1 \\

                2 & 48.5 & 54.1 & 36.0 &  40.4 \\

			4 & 49.2 & 54.6 & 36.8 &  41.3 \\

			8 & \textbf{50.3} & \textbf{55.2} & 37.1 &  42.5 \\

			16 & 50.2 & 55.0 & \textbf{37.4} &  \textbf{42.9} \\
			32 & 49.7 & 54.5 & 37.1 &  42.7 \\
   \bottomrule
			\end{tabular}}
   \vspace{-0.5mm}
   \caption{Ablation study on the effect of learnable text prompts on VGGSound-Single and VGGSound-Duet datasets. Integration of learnable prompts results in noticable performance improvements in both single and multi-source localization.}
	\label{tab:exp_prompt_length}
   \vspace{-1.5em}
\end{table}

\vspace{-0.5em}
\subsection{Use of learnable text prompts}
\vspace{-0.5em}
\label{prompt}
To disentangle multi-source mixtures, we primarily use the class label of each sounding source as text prompts (\textit{e.g.} dog barking, snake hissing). Inspired by the conditional text prompt learning in ~\cite{zhou2022conditional}, we study the use of learnable prompts along with class label representations. Instead of only using class label representation, a set of learnable embedding is integrated with class label embedding that provides additional flexibility to adapt the text prompt to target objective. We analyze the sound source localization performance with different length of learnable prompts. The performances are reported in Table~\ref{tab:exp_prompt_length}.
We notice considerable performance improvements by optimizing learnable prompts in both single and multi-source cases. However, such prompts limit zero-shot use cases to unseen classes.

\vspace{-0.5em}
\section{Conclusion}
\vspace{-0.5em}

In this paper, we propose T-VSL, a novel text-guided multi-source visual sound source localization framework that can disentangle one-to-one audio-visual correspondence from multi-source mixtures.
We leverage the text modality to guide fine-grained feature separation and localization, from noisy audio and visual features of mixtures, by exploiting the joint embedding space of AudioCLIP. 
In comparison with SOTA multi-source baselines, our method shows superior zero-shot transfer across datasets. 
Moreover, our method demonstrates notable robustness on challenging test mixtures with higher number of sources than training scenarios.
In both single-source and multi-source localization, our method largely outperforms existing weakly-supervised and self-supervised baselines on three benchmark datasets.

\subsection*{Acknowledgements}
\vspace{-0.5mm}
This research was supported in part by ONR Minerva program, iMAGiNE - the Intelligent Machine Engineering Consortium at UT Austin, and a UT Cockrell School of Engineering Doctoral Fellowship.

{
    \small
    \bibliographystyle{ieeenat_fullname}
    \bibliography{reference}
}

\appendix
\clearpage
\setcounter{page}{1}

\section{Significant Differences Between Concurrent CLIP-SSL and T-VSL}
\label{diff}
In comparison with the concurrent CLIP-based sound source localization framework CLIP-SSL~\cite{park2023can}, our proposed T-VSL has several significant differences for multi-source localization from mixtures. We highlight the major differences as follows:

1) \textbf{Use of Self-supervised Pre-trained Encoders.} 
CLIP-SSL used an off-the-shelf pre-trained mask-generator that relies on large-scale densely supervised pre-training on image segmentation datasets.
However, sound source localization often demands complex spatio-temporal reasoning across audio and vision modalities, which is absent in image segmentation task.
In contrast, we directly use self-supervised pre-trained AudioCLIP~\cite{guzhov2022audioclip} model as our backbone, and introduced an weakly-supervised sound source localization framework. 
Therefore, our method inherently learns audio-visual correspondence for sound source localization without being limited by large-scale image segmentation supervised pre-training  constraints as CLIP-SSL.

2) \textbf{Text Guidance as Weak Supervision to Noisy Audio and Vision.}
The primary focus of CLIP-SSL is to replace the text query encoder of the supervised baseline with an audio encoder. 
However, environmental audio contains significant noises from background sources in contrast to cleaner text modality. 
In addition, presence of silent objects in visual scenes make the audio-visual correspondence more challening.
Instead of replacing one modality (text with audio), the proposed T-VSL introduces a joint learning across audio, vision, and text modality utilizing the grounded tri-modal embedding space of AudioCLIP.
Our approach particularly leverages the text modality as weak supervision to learn audio-visual correspondence in noisy mixtures, that effectively exploits all three modalities.

3) \textbf{Disentanglement of Multi-Source Mixtures.}
The proposed T-VSL attempts to solve multi-source localization problem in an weakly supervised manner \textit{without having access to single-source audio-visual cues}. 
Since both audio and visual modality contain noises from background sources, it is particularly challenging to learn their correspondence in multi-source scenarios. 
By leveraging the text representation of single-source sounds, we introduce multi-source audio-visual feature disentanglement for enhanced localization performance.
In contrast, CLIP-SSL focuses on directly learning audio-visual correspondence following existing single-source baselines without explicitly tackling the multi-source localization problem.
Such an approach often struggles in learning audio-visual correspondence in challenging multi-source mixtures, when single source sounds and corresponding visual objects are not available.

\begin{figure*}[h]
\centering
\includegraphics[width=0.98\linewidth]{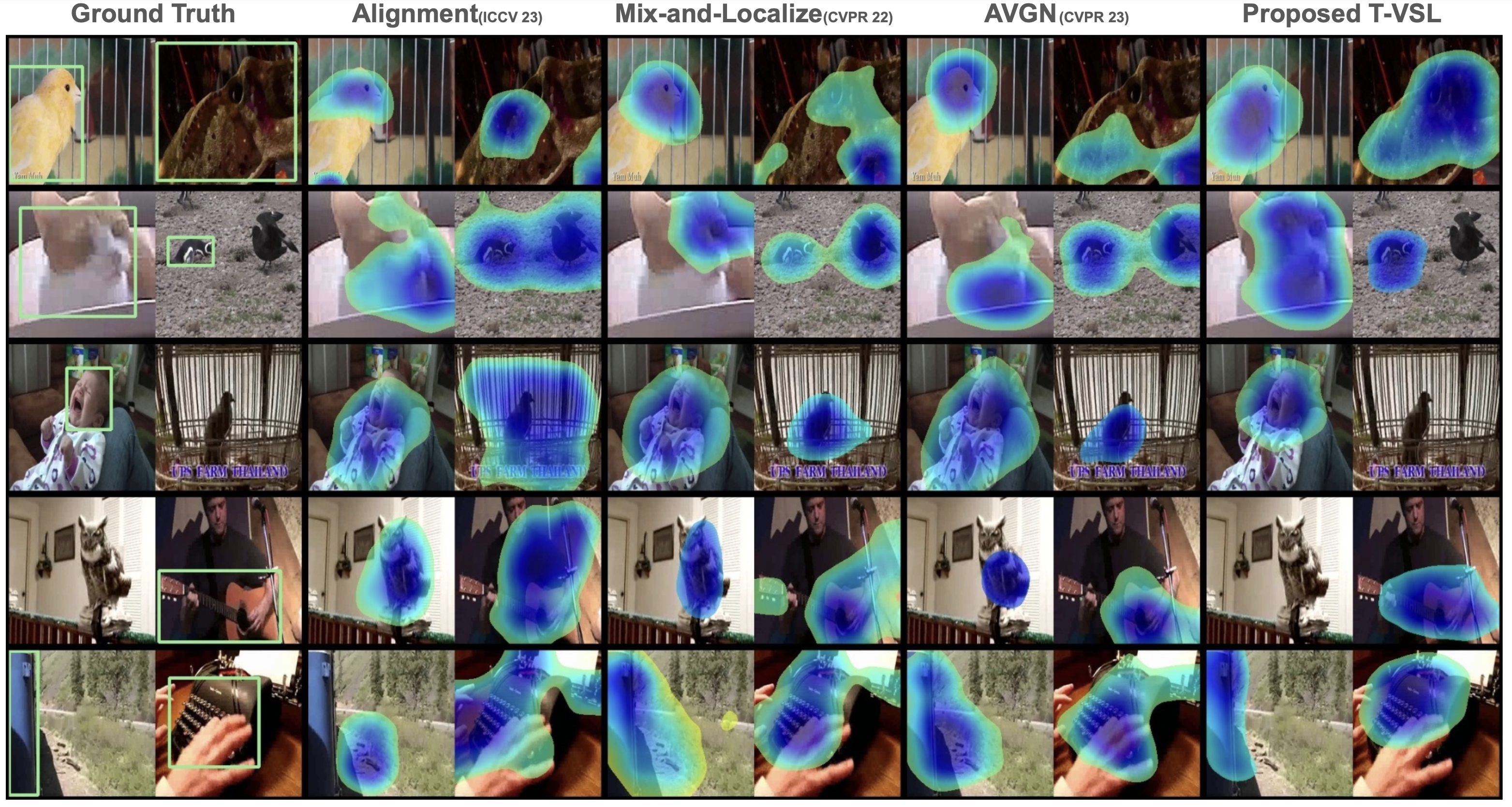}
\caption{
We present additional qualitative comparisons on challenging multi-source localization with SOTA single and multi-source baseline methods. Here, blue color represents high-attention values to the sounding object, and red color represents low-attention values. 
Similar to our prior observation, the proposed T-VSL generates more precise localization maps for sounding sources by selectively isolating the sounding regions from the background.
}
\label{fig: exp_vis_suppl}
\end{figure*}
\begin{table*}[t]
	\renewcommand\tabcolsep{6.0pt}
    \renewcommand{\arraystretch}{1.1}
	\centering
	\scalebox{1.0}{
		\begin{tabular}{ccccccc}
			\toprule
			\multicolumn{1}{c}{\multirow{2}{*}{\begin{tabular}[c]{@{}c@{}} Stage \\ Number \end{tabular}}} & \multicolumn{1}{c}{\multirow{2}{*}{\begin{tabular}[c]{@{}c@{}} Method \end{tabular}}} & \multicolumn{2}{c}{VGGSound-Single} & \multicolumn{2}{c}{VGGSound-Duet} \\
                \cmidrule(lr){3-4} \cmidrule(lr){5-6}
			& & AP(\%) & IoU@0.5(\%) & CAP(\%) & CIoU@0.3(\%)  \\ 	
			\midrule
                Single-Stage & AVGN \scriptsize CVPR23   & 44.1 & 49.6 & 31.9 &  37.8 \\

                Single-Stage & T-VSL  & 46.8 & 51.5 & 33.7 &  38.7 \\

			Two-Stage & T-VSL  & \textbf{48.1} & \textbf{53.7} & \textbf{35.7} &  \textbf{40.1} \\

			\bottomrule
			\end{tabular}}
   \vspace{-0.5em}
   \caption{Quantitative comparison on single and multi-stage architectures on VGGSound-Single and VGGSound-Duet datasets. Same AudioCLIP encoders are used. Proposed two-stage method generates superior performance compared to its single-stage counterpart. }
	\label{tab:stage}
\end{table*}

\section{Comparison between one-stage and two-stage architectures in T-VSL}
\label{onestage}
We introduce a two-stage architecture in the proposed T-VSL, that consists of audio-visual class instance detection followed by iterative localization of each sounding source present in the multi-source mixture.
In general, this two-stage approach simplifies the problem of challenging multi-source localization by initially detecting common audio-visual instances present in both audio and visual modality. 
However, we also study an one-stage alternative of the proposed T-VSL following AVGN~\cite{mo2023audio},  by removing the audio-visual class instance detection stage, and by incorporating all $N$-class text embedding of sounding sources in both audio and visual conditioning blocks, without explicitly separating $K (K \leq N)$ class instances.
We present the quantitative comparisons on VGGSound-Single and VGGSound-Duet datasets in Table~\ref{tab:stage}.
We note that the two-stage method  in T-VSL  achieves $+2.2$ IoU@0.5\% and $+1.4$ CIoU@0.3\% improvements on VGGSound-Single and VGGSound-Duet datasets, respectively, compared to its single-stage counterpart.
We hypothesize that the proposed two-stage method greatly reduces the effect of background noises in challenging audio-visual correspondence learning from natural mixtures. 
In contrast, single-stage method introduces additional noises from background sources in audio-visual conditioning blocks for not explicitly suppressing background conditions.

\section{Qualitative Comparisons}
\label{qual}
We present additional qualitative comparisons of the proposed T-VSL with SOTA methods on challenging multi-source localization in Figure~\ref{fig: exp_vis_suppl}. 
We note that the proposed T-VSL consistently achieves superior performance by generating more precise localization maps compared to other SOTA baselines, which follows our prior observation.
In addition, T-VSL can selectively isolate the non-sounding silent objects, where other baselines struggle in isolating the silent objects present in the surroundings.
These qualitative results demonstrate the effectiveness of T-VSL in disentangling multi-source mixtures as well as in learning audio-visual correspondence for sound source localization.

\end{document}